\documentclass[runningheads]{llncs}
\usepackage[T1]{fontenc}
\usepackage{graphicx}
\usepackage{hyperref}
\usepackage[numbers,sort&compress]{natbib}

\usepackage{mathrsfs}
\usepackage{amsmath}
\usepackage{amsfonts}
\usepackage{booktabs}
\usepackage{multirow}
\usepackage{xurl} 
\hypersetup{
    breaklinks=true 
}

\begin{document}
\title{PictSure: Pretraining Embeddings Matters for In-Context Learning Image Classifiers}
\titlerunning{PictSure: Pretraining Embeddings for ICL Image Classifiers}
%

\author{Lukas Schiesser\thanks{These authors contributed equally to this work.}\inst{1} \and
Cornelius Wolff\protect\footnotemark[1]\inst{2} \and
Sophie Haas\inst{1} \and
Simon Pukrop\inst{1}}
\authorrunning{L. Schiesser et al.}
%
\institute{German Research Center for AI (DFKI), Hamburger Str. 24, 49084 Osnabrück, Germany\\
\email{\{lukas.schiesser, sophie.haas, simon.pukrop\}@dfki.de} \and
Centrum Wiskunde \& Informatica (CWI), Science Park 123, 1098 XG Amsterdam, Netherlands\\
\email{cornelius.wolff@cwi.nl}}

\maketitle              
\begin{abstract}
    Building image classification models remains cumbersome in data-scarce domains, where collecting large labeled datasets is impractical. In-context learning (ICL) is a promising paradigm for few-shot image classification (FSIC), but prior work has underexplored the relative importance of encoder pretraining versus fusion-layer training data. We present PictSure, a vision-only ICL family of models that demonstrates the potential of easy-to-use fusion transformer architectures, as well as the need for better embedding representations across a wider range of image domains. In both in-domain and out-of-domain evaluations, we find that representation quality induced by pretraining strongly correlates with downstream ICL performance. Crucially, varying the training dataset for the fusion transformer, from ImageNet alone to diverse multi-domain mixtures, provides limited additional performance gains under the evaluated settings, demonstrating that the fusion layer appears capable of adapting effectively once embeddings are sufficiently structured. These results show that the bottleneck in visual ICL is representation quality, not fusion-module training diversity. To facilitate adoption and reproducibility, we release all model weights as open-source artifacts and provide an MCP server that exposes PictSure as a callable tool for LLM-based agentic systems, enabling few-shot image classification to be invoked directly within AI pipelines without integration overhead. Code can be found at \href{https://github.com/PictSure}{\url{https://github.com/PictSure}} and models at \href{https://huggingface.co/pictsure}{\url{https://huggingface.co/pictsure}}.
\keywords{In-Context Learning \and Few-Shot Image Classification \and Meta Learning \and Domain Generalization}
\end{abstract}
\section{Introduction}
Collecting large labeled datasets is impractical in many real-world domains such as medical imaging or agriculture, where expert annotation is costly and privacy constraints may apply~\cite{zhang_challenges_2024, s_privacy-preserving_2024, yang2022survey, zhou2024few}. Few-shot image classification (FSIC) addresses this by enabling models to recognize unseen classes from only a handful of examples~\cite{snell_prototypical_2017}. In-context learning (ICL) is a particularly attractive paradigm for FSIC: a transformer is provided with labeled support image-label pairs alongside a query, and predicts the query label without any parameter updates at inference time~\cite{brown_language_2020, fifty_context-aware_2023}. This eliminates costly fine-tuning and supports immediate deployment across changing task definitions.

Existing visual ICL work, such as CAML~\cite{fifty_context-aware_2023} and SgVA-CLIP~\cite{peng_sgva-clip_2024}, rely on CLIP-based encoders whose language-grounded representations struggle with fine-grained or domain-shifted categories~\cite{radford_learning_2021}. Prior work has largely overlooked a more fundamental question: \textit{how much does the choice of encoder and its pretraining strategy determine ICL performance?} Evidence that a linear probe on strong embeddings can outperform sophisticated meta-learners~\cite{tian_rethinking_2020} suggests representation quality may be the dominant factor.

We present \textbf{PictSure}, a set of vision-only ICL models designed to tackle this question. To isolate the effect of image encoding on the downstream ICL task, we keep the fusion architecture fixed and systematically vary the encoder and pretraining objective, comparing a supervised ResNet~\cite{he_deep_2016} pretrained on ImageNet-21k~\cite{ridnik_imagenet-21k_2021}, a contrastive CLIP encoder~\cite{radford_learning_2021}, and a self-supervised DINOv2~\cite{oquab_dinov2_nodate}. Our central findings are twofold. First, the encoder and its pretraining matter far more than the training set of the ICL transformer itself: models without pretrained encoders fail to converge, while a well-pretrained encoder enables strong in-domain and out-of-domain transfer. Second, expanding the number and diversity of training datasets for the fusion transformer yields negligible gains. The fusion layer learns to read the embedding space from ImageNet alone, demonstrating how readily the ICL transformer adapts once representations are well-structured. This confirms that the bottleneck is not the fusion module's exposure to diverse tasks, but the quality of the embedding space, suggesting that better representations that improve class separability are a key lever for performance gains. The main contributions are: (1)~PictSure, a transformer-based vision-only in-context learner requiring no fine-tuning or language supervision; (2)~a systematic study showing encoder pretraining quality is the dominant performance driver in visual ICL; (3)~competitive out-of-domain results on medical benchmarks (Brain Tumor, OrganCMNIST), demonstrating the advantage of purely visual embeddings for domain generalization; (4)~a fully open-source baseline for future FSIC research, providing a clean, reproducible foundation for studying transformer-based fusion layers under varying encoder designs, including an MCP server that exposes PictSure as a tool for agentic systems to perform few-shot image classification directly within AI pipelines.

\section{Background \& Model}
\label{sec:model_overview}

\noindent\textbf{FSIC and ICL.}
An alternative paradigm is ICL which eliminates parameter updates by conditioning on a demonstration set $\mathscr{D} = \{(x_i, y_i)\}_{i=1}^{m}$ provided at inference time to predict the label of a query~\citep{brown_language_2020}:
\[F_\theta: (\mathscr{D}, x_q) \rightarrow \hat{y_q}\]
Prior work such as CAML~\citep{fifty_context-aware_2023} extends this paradigm to vision by leveraging CLIP-based encoders to perform ICL over images. However, these approaches inherit a key limitation of language-aligned representations: they often fail to preserve fine-grained visual distinctions and degrade under domain shift, as textual descriptions provide only weak supervision for subtle inter-class differences~\cite{radford_learning_2021, fifty_context-aware_2023}.

\begin{figure}[h]
\centering
\includegraphics[width=.8\linewidth]{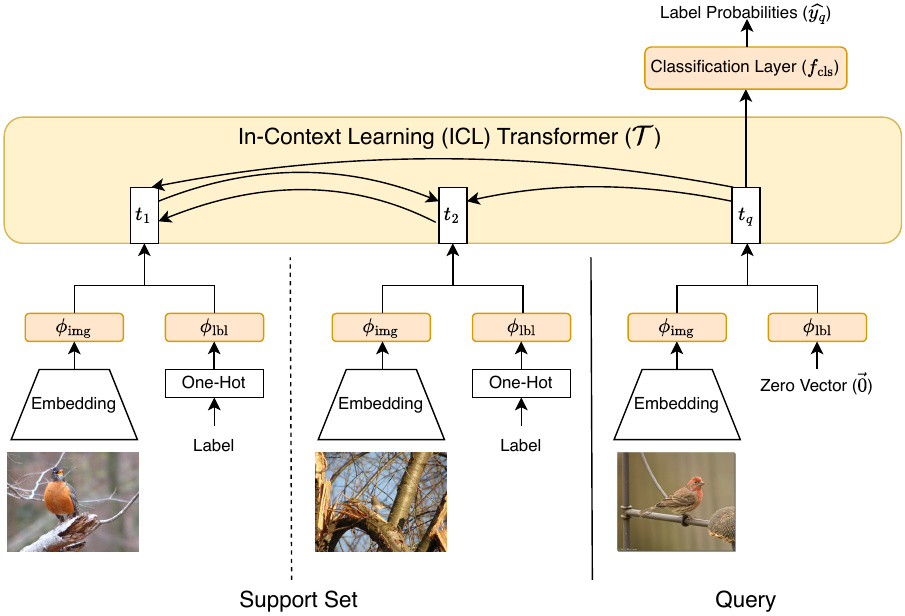}
\caption{Overview of the PictSure architecture}
\label{fig:architecture}
\end{figure}

\noindent\textbf{PictSure Architecture.}
Each support image $x_i$ is encoded by a frozen visual encoder $\phi_{\text{img}}$ into $v_i \in \mathbb{R}^d$; its one-hot label $y_i$ is projected by $\phi_{\text{lbl}}$ into $\ell_i \in \mathbb{R}^d$. These are concatenated into a joint token $t_i = [v_i;\, \ell_i]$; the query uses $t_q = [v_q;\, \vec{0}]$. The ICL transformer $\mathcal{T}$ then processes the full sequence under an asymmetric mask: support tokens attend to each other freely, the query attends to all support tokens, but support tokens cannot attend to the query---mirroring meta-inference without mutual contamination. The output query representation is classified via:
\[
\hat{y_q} = f_{\text{cls}}\!\left(\mathcal{T}([t_1, \dots, t_m, t_q])\right) \in \mathbb{R}^n
\]
The transformer uses four blocks, eight attention heads, and model dimension 1024. The broad architecture is illustrated in  Figure~\ref{fig:architecture}. Models are trained on 10-way 5-shot episodes sampled from ImageNet-21K~\citep{ridnik_imagenet-21k_2021} (10{,}000 episodes per epoch, images at $224{\times}224$).

\section{Training}
\label{sec:training}

\noindent We evaluate three embedding backbones integrated into PictSure, varying only the encoder (with frozen weights) and projection while keeping the ICL transformer architecture the same.
ResNet \citep{he_deep_2016} is a supervised convolutional network trained on ImageNet1K using standard classification objectives.
CLIP \citep{radford_learning_2021} is a Vision Transformer trained via contrastive alignment on 400 million image--text pairs, with its image encoder frozen during integration.
DINOv2 \citep{oquab_dinov2_nodate} is a ViT trained through self-distillation and masked-image modeling on 142 million images without any labels.
Across all settings, pretrained and frozen encoders prove necessary for consistent convergence and strong few-shot performance.

We study two fusion-transformer training configurations that vary only in the composition of the training data.
The \textit{ImageNet} setting trains exclusively on ImageNet-21K~\citep{ridnik_imagenet-21k_2021}, the standard pretraining corpus used in prior work.
\begin{sloppypar}
The \textit{Default} setting replaces this with a multi-domain mixture of sixteen datasets spanning diverse visual categories: ILSVRC-2012~\citep{russakovsky_imagenet_2015} (20\%), Places365~\citep{zhou_places_2018} (10\%), VGGFace2~\citep{cao_vggface2_2018} (10\%), Products10K~\citep{bai_products-10k_2020} (10\%), LEGO Parts~\citep{noauthor_lego_2025} (10\%), Food-101~\citep{fleet_food-101_2014} (5\%), SD-198~\citep{sun2016benchmark} (5\%), ChestX~\citep{wang_chestx-ray8_2017} (5\%), Cars196~\citep{krause_3d_2013} (5\%), WildlifeReID-10K~\citep{adam_wildlifereid-10k_2025} (5\%), EuroSAT~\citep{helber_eurosat_2019} (4\%), PlantVillage (2.5\%), Fruits-360~\citep{sa2016deepfruits} (2.5\%), HAM10000~\citep{tschandl_ham10000_2018} (2\%), Oxford Flowers~\citep{nilsback_automated_2008} (2\%), and KAU-BCMD~\citep{alsolami_king_2021} (2\%).
This mixture deliberately spans general, fine-grained, medical, agricultural, and remote-sensing imagery to test whether broader domain coverage during fusion training improves out-of-domain generalization.
\end{sloppypar}

\section{Model Evaluation}
\label{sec:evaluation}
We evaluated the performance of our PictSure model variations across a diverse range of visual domains. Both in-domain and out-of-domain datasets, comprising general, agricultural, and medical imagery were used for the evaluation, namely tieredImageNet \cite{ren_meta-learning_2018},  PlantDoc \cite{singh_plantdoc_2020},  Brain Tumor Classification (MRI) dataset \cite{chakrabarty_brain_nodate}, and OCTMNIST \cite{yang_medmnist_2023}. All evaluations use 5-way $k$-shot tasks with $k \in \{1, 5\}$, averaged over 2000 episodes per dataset and reported as mean accuracy $\pm$ standard error.

\begin{figure}[h]
\centering
\includegraphics[width=\linewidth]{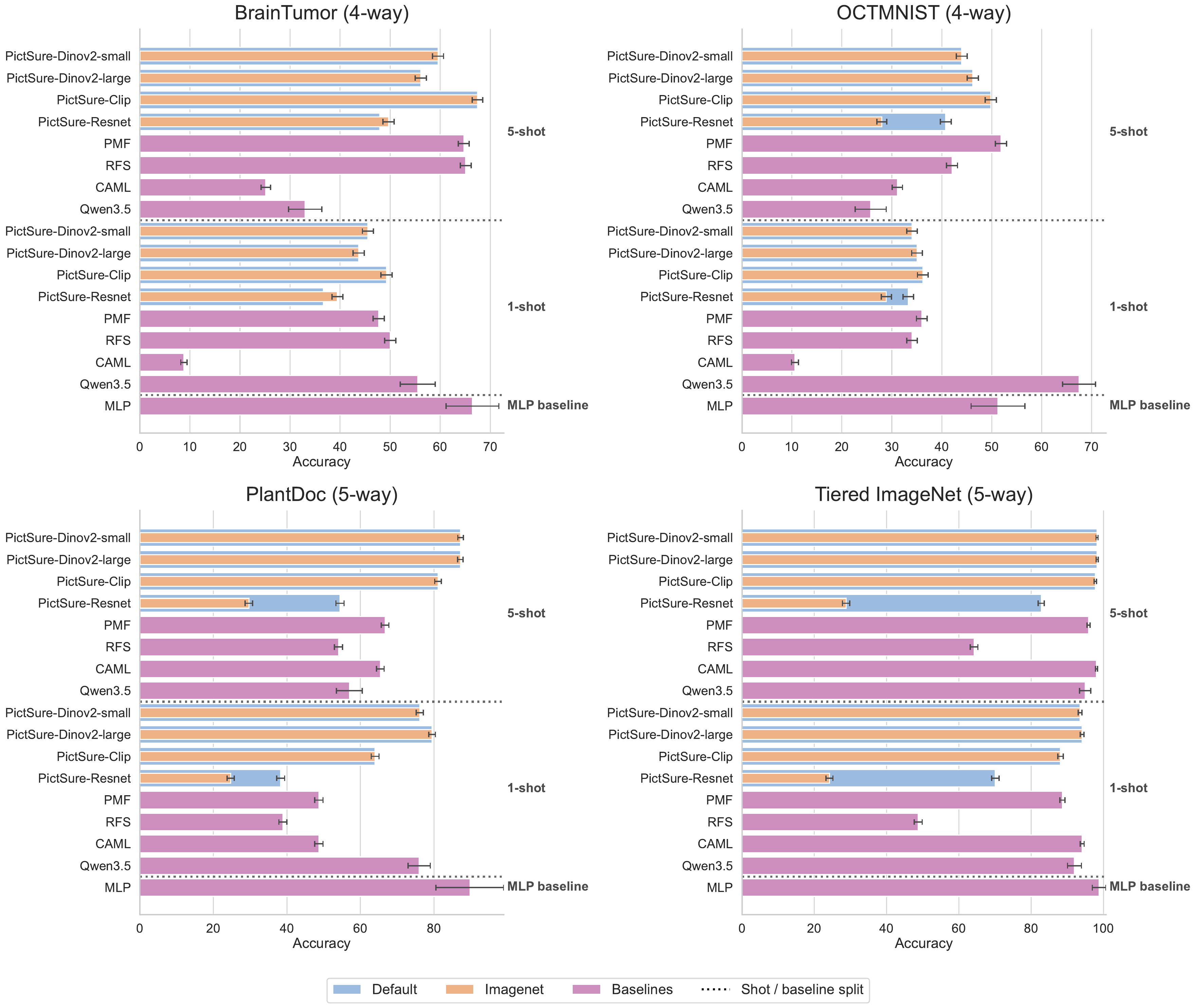}
\caption{Test results across all PictSure model variants compared to baselines}
\label{fig:results}
\end{figure}

\noindent\textbf{Baselines.}
We compare multiple variants of PictSure against several established baselines. First, we include the representation learning baseline of \citet{tian_rethinking_2020}, consisting of a pretrained embedding network followed by a task-specific linear classifier trained on support embeddings (RFS). Second, we consider the method of \citet{hu_pushing_2022}, which combines large-scale pretraining, ProtoNet-based meta-training, and task-specific fine-tuning (PMF).
We further include CAML~\citep{fifty_context-aware_2023}, a CLIP-based approach to few-shot learning. In addition, we evaluate the multimodal model Qwen3.5 122B in a pure ICL setting, where the model receives $m$ support image–label pairs and a query (images resized to $224 \times 224$) without parameter updates.
All methods are evaluated on 5-way 1-shot and 5-way 5-shot tasks. We report mean accuracy and standard error across episodes. Figure~\ref{fig:results} summarizes the results.
As a supervised baseline, we train a multilayer perceptron (MLP) directly on the pre-computed embeddings.
The network consists of an input projection layer followed by four residual blocks, each comprising two
fully-connected layers with batch normalization, ReLU activations, and dropout, with a linear output head.
To match the evaluation protocol of our primary method, we repeatedly sample random subsets of 10 classes
from each dataset (200 runs per dataset) and train a separate MLP on each subset. Models are optimized with AdamW using a linear learning rate decay schedule and early stopping.

\section{Analysis}
\label{sec:discussion}
Encoder quality appears to be one of the strongest factors influencing PictSure’s performance. DINOv2 and CLIP variants consistently outperform the ResNet-based variant across all benchmarks, most clearly on tieredImageNet and PlantDoc in both shot settings. These results support the hypothesis that strong representations may contribute more to performance than increased adaptation complexity in the evaluated setting.
The fusion transformer showed limited sensitivity to training-data composition across the evaluated datasets. The \textit{Default} and \textit{Imagenet} settings yield nearly identical results in our experiments, suggesting ImageNet-21K-style supervision is sufficient for learning support–query interaction and that further gains are more strongly associated with representation quality than data diversity. The exception is ResNet-50, which shows a modest sensitivity to training data composition. We hypothesize that this is because ResNet's global-average-pooled representations carry less structural information than ViT patch tokens, leaving the fusion module with a less separable embedding space to work from; in that regime, exposure to more varied training tasks may partially compensate for the representational deficit, whereas with a strong encoder the fusion layer benefits less from additional diversity.
PictSure's relative performance depends on the target domain. It is highly competitive on tieredImageNet and particularly strong on PlantDoc, where DINOv2 variants outperform RFS, PMF, and CAML. On OCTMNIST and Brain Tumor it remains competitive, though PMF, RFS, and Qwen3.5 can match or exceed it. PictSure demonstrates competitive cross-domain transfer performance, with its clearest advantages on natural-image benchmarks.
With CLIP embeddings frozen, PictSure remains relatively consistent across domains, whereas CAML shows a clear drop on the medical datasets, despite using the same encoder models. Thus, the main difference between adaptation methods appears to be robustness to domain shift rather than absolute performance gains. The supervised MLP baseline reinforces this: its strong performance confirms that pretrained embeddings already contain substantial task-relevant information.
Beyond accuracy, PictSure offers a favorable efficiency profile, substantially smaller than CAML and Qwen3.5 while remaining competitive across benchmarks, with advantages over RFS and PMF in transfer performance, lightweight fusion, and stable inference.

\section{Outlook and Conclusion}
\label{sec:conclusion}

Our results support an encoder-centric perspective on visual ICL design. The fusion transformer learns quickly and reliably once embeddings are well-structured, while changes in fusion complexity or training data provide only limited gains when representations are weak. Representation quality appears to be the limiting factor, not the transformer's ability to perform support-query fusion.

This points to a clear path forward: if ICL transformers already act as robust readers of embedding spaces, further gains are to be found in building better visual representations through stronger pretraining objectives, better encoder architectures, and improved encoder-task alignment under distribution shift.

To lower the barrier to adoption, we release an MCP server exposing PictSure as a callable tool for agentic systems, enabling few-shot image classification within AI pipelines without integration overhead. Future work should prioritize representation learning at scale while keeping the fusion module lightweight, including improved pretraining curricula, broader encoder families, and extended class coverage beyond the current 10-way setup.

%
%
%
\bibliographystyle{splncs04nat}
\bibliography{PictSure}

@inproceedings{singh_plantdoc_2020,
	address = {Hyderabad India},
	title = {{PlantDoc}: {A} {Dataset} for {Visual} {Plant} {Disease} {Detection}},
	isbn = {978-1-4503-7738-6},
	shorttitle = {{PlantDoc}},
	url = {https://dl.acm.org/doi/10.1145/3371158.3371196},
	doi = {10.1145/3371158.3371196},
	language = {en},
	urldate = {2025-02-16},
	booktitle = {Proceedings of the 7th {ACM} {IKDD} {CoDS} and 25th {COMAD}},
	publisher = {ACM},
	author = {Singh, Davinder and Jain, Naman and Jain, Pranjali and Kayal, Pratik and Kumawat, Sudhakar and Batra, Nipun},
	month = jan,
	year = {2020},
	pages = {249--253},
}

@inproceedings{wang_chestx-ray8_2017,
	title = {{ChestX}-ray8: {Hospital}-{Scale} {Chest} {X}-{Ray} {Database} and {Benchmarks} on {Weakly}-{Supervised} {Classification} and {Localization} of {Common} {Thorax} {Diseases}},
	language = {en},
	booktitle = {2017 {IEEE} {Conference} on {Computer} {Vision} and {Pattern} {Recognition} ({CVPR})},
	author = {Wang, Xiaosong and Peng, Yifan and Lu, Le and Lu, Zhiyong and Bagheri, Mohammadhadi and Summers, Ronald M},
	year = {2017},
	pages = {3462 -- 3471},
}

@article{s_privacy-preserving_2024,
	title = {Privacy-{Preserving} {Breast} {Cancer} {Classification}: {A} {Federated} {Transfer} {Learning} {Approach}},
	volume = {37},
	issn = {2948-2933},
	shorttitle = {Privacy-{Preserving} {Breast} {Cancer} {Classification}},
	url = {https://link.springer.com/10.1007/s10278-024-01035-8},
	doi = {10.1007/s10278-024-01035-8},
	language = {en},
	number = {4},
	urldate = {2025-02-16},
	journal = {Journal of Imaging Informatics in Medicine},
	author = {S, Selvakanmani and Dharani Devi, G and V, Rekha and Jeyalakshmi, J},
	month = feb,
	year = {2024},
	pages = {1488--1504},
}

@article{yang_medmnist_2023,
	title = {{MedMNIST} v2 - {A} large-scale lightweight benchmark for {2D} and {3D} biomedical image classification},
	volume = {10},
	issn = {2052-4463},
	url = {https://www.nature.com/articles/s41597-022-01721-8},
	doi = {10.1038/s41597-022-01721-8},
	language = {en},
	number = {1},
	urldate = {2025-02-16},
	journal = {Scientific Data},
	author = {Yang, Jiancheng and Shi, Rui and Wei, Donglai and Liu, Zequan and Zhao, Lin and Ke, Bilian and Pfister, Hanspeter and Ni, Bingbing},
	month = jan,
	year = {2023},
	pages = {41},
}

@inproceedings{fifty_context-aware_2023,
	title = {Context-{Aware} {Meta}-{Learning}},
	url = {https://openreview.net/forum?id=lJYAkDVnRU},
	language = {en},
	urldate = {2025-03-31},
	author = {Fifty, Christopher and Duan, Dennis and Junkins, Ronald Guenther and Amid, Ehsan and Leskovec, Jure and Re, Christopher and Thrun, Sebastian},
	month = oct,
	year = {2023},
}

@inproceedings{brown_language_2020,
	title = {Language {Models} are {Few}-{Shot} {Learners}},
	volume = {33},
	url = {https://papers.nips.cc/paper/2020/hash/1457c0d6bfcb4967418bfb8ac142f64a-Abstract.html},
	urldate = {2025-02-27},
	booktitle = {Advances in {Neural} {Information} {Processing} {Systems}},
	publisher = {Curran Associates, Inc.},
	author = {Brown, Tom and Mann, Benjamin and Ryder, Nick and Subbiah, Melanie and Kaplan, Jared D and Dhariwal, Prafulla and Neelakantan, Arvind and Shyam, Pranav and Sastry, Girish and Askell, Amanda and Agarwal, Sandhini and Herbert-Voss, Ariel and Krueger, Gretchen and Henighan, Tom and Child, Rewon and Ramesh, Aditya and Ziegler, Daniel and Wu, Jeffrey and Winter, Clemens and Hesse, Chris and Chen, Mark and Sigler, Eric and Litwin, Mateusz and Gray, Scott and Chess, Benjamin and Clark, Jack and Berner, Christopher and McCandlish, Sam and Radford, Alec and Sutskever, Ilya and Amodei, Dario},
	year = {2020},
	pages = {1877--1901},
}

@article{ren_meta-learning_2018,
	title = {{META}-{LEARNING} {FOR} {SEMI}-{SUPERVISED} {FEW}-{SHOT} {CLASSIFICATION}},
	language = {en},
	author = {Ren, Mengye and Triantaﬁllou, Eleni and Ravi, Sachin and Snell, Jake and Swersky, Kevin and Tenenbaum, Joshua B and Larochelle, Hugo and Zemel, Richard S},
	year = {2018},
}

@inproceedings{hu_pushing_2022,
	address = {New Orleans, LA, USA},
	title = {Pushing the {Limits} of {Simple} {Pipelines} for {Few}-{Shot} {Learning}: {External} {Data} and {Fine}-{Tuning} {Make} a {Difference}},
	copyright = {https://doi.org/10.15223/policy-029},
	isbn = {978-1-6654-6946-3},
	shorttitle = {Pushing the {Limits} of {Simple} {Pipelines} for {Few}-{Shot} {Learning}},
	url = {https://ieeexplore.ieee.org/document/9879354/},
	doi = {10.1109/CVPR52688.2022.00886},
	language = {en},
	urldate = {2025-04-23},
	booktitle = {2022 {IEEE}/{CVF} {Conference} on {Computer} {Vision} and {Pattern} {Recognition} ({CVPR})},
	publisher = {IEEE},
	author = {Hu, Shell Xu and Li, Da and Stuhmer, Jan and Kim, Minyoung and Hospedales, Timothy M.},
	month = jun,
	year = {2022},
	pages = {9058--9067},
}

@inproceedings{ridnik_imagenet-21k_2021,
	title = {{ImageNet}-{21K} {Pretraining} for the {Masses}},
	volume = {1},
	url = {https://datasets-benchmarks-proceedings.neurips.cc/paper_files/paper/2021/file/98f13708210194c475687be6106a3b84-Paper-round1.pdf},
	booktitle = {Proceedings of the {Neural} {Information} {Processing} {Systems} {Track} on {Datasets} and {Benchmarks}},
	author = {Ridnik, Tal and Ben-Baruch, Emanuel and Noy, Asaf and Zelnik, Lihi},
	editor = {Vanschoren, J. and Yeung, S.},
	year = {2021},
}

@article{snell_prototypical_2017,
	title = {Prototypical networks for few-shot learning},
	volume = {30},
	journal = {Advances in neural information processing systems},
	author = {Snell, Jake and Swersky, Kevin and Zemel, Richard},
	year = {2017},
}

@inproceedings{radford_learning_2021,
	title = {Learning transferable visual models from natural language supervision},
	booktitle = {International conference on machine learning},
	publisher = {PmLR},
	author = {Radford, Alec and Kim, Jong Wook and Hallacy, Chris and Ramesh, Aditya and Goh, Gabriel and Agarwal, Sandhini and Sastry, Girish and Askell, Amanda and Mishkin, Pamela and Clark, Jack and {others}},
	year = {2021},
	pages = {8748--8763},
}

@article{peng_sgva-clip_2024,
	title = {{SgVA}-{CLIP}: {Semantic}-{Guided} {Visual} {Adapting} of {Vision}-{Language} {Models} for {Few}-{Shot} {Image} {Classification}},
	volume = {26},
	copyright = {https://ieeexplore.ieee.org/Xplorehelp/downloads/license-information/IEEE.html},
	issn = {1520-9210, 1941-0077},
	shorttitle = {{SgVA}-{CLIP}},
	url = {https://ieeexplore.ieee.org/document/10243119/},
	doi = {10.1109/TMM.2023.3311646},
	language = {en},
	urldate = {2025-05-13},
	journal = {IEEE Transactions on Multimedia},
	author = {Peng, Fang and Yang, Xiaoshan and Xiao, Linhui and Wang, Yaowei and Xu, Changsheng},
	year = {2024},
	pages = {3469--3480},
}

@inproceedings{he_deep_2016,
	address = {Las Vegas, NV, USA},
	title = {Deep {Residual} {Learning} for {Image} {Recognition}},
	isbn = {978-1-4673-8851-1},
	url = {http://ieeexplore.ieee.org/document/7780459/},
	doi = {10.1109/CVPR.2016.90},
	language = {en},
	urldate = {2025-05-13},
	booktitle = {2016 {IEEE} {Conference} on {Computer} {Vision} and {Pattern} {Recognition} ({CVPR})},
	publisher = {IEEE},
	author = {He, Kaiming and Zhang, Xiangyu and Ren, Shaoqing and Sun, Jian},
	month = jun,
	year = {2016},
	pages = {770--778},
}

@article{zhang_challenges_2024,
	title = {On the challenges and perspectives of foundation models for medical image analysis},
	volume = {91},
	journal = {Medical image analysis},
	publisher = {Elsevier},
	author = {Zhang, Shaoting and Metaxas, Dimitris},
	year = {2024},
	pages = {102996},
}

@inproceedings{tian_rethinking_2020,
	title = {Rethinking few-shot image classification: a good embedding is all you need?},
	booktitle = {Computer {Vision}–{ECCV} 2020: 16th {European} {Conference}, {Glasgow}, {UK}, {August} 23–28, 2020, {Proceedings}, {Part} {XIV} 16},
	publisher = {Springer},
	author = {Tian, Yonglong and Wang, Yue and Krishnan, Dilip and Tenenbaum, Joshua B and Isola, Phillip},
	year = {2020},
	pages = {266--282},
}

@misc{chakrabarty_brain_nodate,
	title={Brain Tumor MRI Dataset},
	url={https://www.kaggle.com/dsv/14832123},
	DOI={10.34740/KAGGLE/DSV/14832123},
	publisher={Kaggle},
	author={Msoud Nickparvar},
	year={2026}
}

@article{russakovsky_imagenet_2015,
	title = {{ImageNet} {Large} {Scale} {Visual} {Recognition} {Challenge}},
	volume = {115},
	issn = {0920-5691, 1573-1405},
	url = {http://link.springer.com/10.1007/s11263-015-0816-y},
	doi = {10.1007/s11263-015-0816-y},
	language = {en},
	number = {3},
	urldate = {2026-01-08},
	journal = {International Journal of Computer Vision},
	author = {Russakovsky, Olga and Deng, Jia and Su, Hao and Krause, Jonathan and Satheesh, Sanjeev and Ma, Sean and Huang, Zhiheng and Karpathy, Andrej and Khosla, Aditya and Bernstein, Michael and Berg, Alexander C. and Fei-Fei, Li},
	month = dec,
	year = {2015},
	pages = {211--252},
}

@article{zhou_places_2018,
	title = {Places: {A} 10 {Million} {Image} {Database} for {Scene} {Recognition}},
	volume = {40},
	copyright = {https://ieeexplore.ieee.org/Xplorehelp/downloads/license-information/IEEE.html},
	issn = {0162-8828, 2160-9292, 1939-3539},
	shorttitle = {Places},
	url = {https://ieeexplore.ieee.org/document/7968387/},
	doi = {10.1109/TPAMI.2017.2723009},
	language = {en},
	number = {6},
	urldate = {2026-01-08},
	journal = {IEEE Transactions on Pattern Analysis and Machine Intelligence},
	author = {Zhou, Bolei and Lapedriza, Agata and Khosla, Aditya and Oliva, Aude and Torralba, Antonio},
	month = jun,
	year = {2018},
	pages = {1452--1464},
}

@incollection{fleet_food-101_2014,
	address = {Cham},
	title = {Food-101 – {Mining} {Discriminative} {Components} with {Random} {Forests}},
	volume = {8694},
	copyright = {http://www.springer.com/tdm},
	isbn = {978-3-319-10598-7 978-3-319-10599-4},
	url = {http://link.springer.com/10.1007/978-3-319-10599-4_29},
	doi = {10.1007/978-3-319-10599-4_29},
	language = {en},
	urldate = {2026-01-08},
	booktitle = {Computer {Vision} – {ECCV} 2014},
	publisher = {Springer International Publishing},
	author = {Bossard, Lukas and Guillaumin, Matthieu and Van Gool, Luc},
	editor = {Fleet, David and Pajdla, Tomas and Schiele, Bernt and Tuytelaars, Tinne},
	year = {2014},
	note = {Series Title: Lecture Notes in Computer Science},
	pages = {446--461},
}

@article{helber_eurosat_2019,
	title = {{EuroSAT}: {A} {Novel} {Dataset} and {Deep} {Learning} {Benchmark} for {Land} {Use} and {Land} {Cover} {Classification}},
	volume = {12},
	copyright = {https://ieeexplore.ieee.org/Xplorehelp/downloads/license-information/IEEE.html},
	issn = {1939-1404, 2151-1535},
	shorttitle = {{EuroSAT}},
	url = {https://ieeexplore.ieee.org/document/8736785/},
	doi = {10.1109/JSTARS.2019.2918242},
	language = {en},
	number = {7},
	urldate = {2026-01-08},
	journal = {IEEE Journal of Selected Topics in Applied Earth Observations and Remote Sensing},
	author = {Helber, Patrick and Bischke, Benjamin and Dengel, Andreas and Borth, Damian},
	month = jul,
	year = {2019},
	pages = {2217--2226},
}

@article{tschandl_ham10000_2018,
	title = {The {HAM10000} dataset, a large collection of multi-source dermatoscopic images of common pigmented skin lesions},
	volume = {5},
	issn = {2052-4463},
	url = {https://www.nature.com/articles/sdata2018161},
	doi = {10.1038/sdata.2018.161},
	language = {en},
	number = {1},
	urldate = {2026-01-08},
	journal = {Scientific Data},
	author = {Tschandl, Philipp and Rosendahl, Cliff and Kittler, Harald},
	month = aug,
	year = {2018},
	pages = {180161},
}

@misc{cao_vggface2_2018,
	title = {{VGGFace2}: {A} dataset for recognising faces across pose and age},
	shorttitle = {{VGGFace2}},
	url = {http://arxiv.org/abs/1710.08092},
	doi = {10.48550/arXiv.1710.08092},
	language = {en},
	urldate = {2026-01-08},
	publisher = {arXiv},
	author = {Cao, Qiong and Shen, Li and Xie, Weidi and Parkhi, Omkar M. and Zisserman, Andrew},
	month = may,
	year = {2018},
	note = {arXiv:1710.08092 [cs]},
}

@inproceedings{nilsback_automated_2008,
	address = {Bhubaneswar, India},
	title = {Automated {Flower} {Classification} over a {Large} {Number} of {Classes}},
	url = {http://ieeexplore.ieee.org/document/4756141/},
	doi = {10.1109/ICVGIP.2008.47},
	language = {en},
	urldate = {2026-01-08},
	booktitle = {2008 {Sixth} {Indian} {Conference} on {Computer} {Vision}, {Graphics} \& {Image} {Processing}},
	publisher = {IEEE},
	author = {Nilsback, Maria-Elena and Zisserman, Andrew},
	month = dec,
	year = {2008},
	pages = {722--729},
}

@misc{noauthor_lego_2025,
	address = {Kaggle},
	title = {Lego {Parts} {Classification} {Dataset} - 2.{4M} {Renders}},
	url = {https://www.kaggle.com/datasets/mihahamyt/lego-parts-classification-dataset-2-4m-renders},
	urldate = {2026-01-10},
	month = oct,
	year = {2025},
}

@misc{adam_wildlifereid-10k_2025,
	title = {{WildlifeReID}-10k: {Wildlife} re-identification dataset with 10k individual animals},
	shorttitle = {{WildlifeReID}-10k},
	url = {http://arxiv.org/abs/2406.09211},
	doi = {10.48550/arXiv.2406.09211},
	language = {en},
	urldate = {2026-01-13},
	publisher = {arXiv},
	author = {Adam, Lukáš and Čermák, Vojtěch and Papafitsoros, Kostas and Picek, Lukas},
	month = apr,
	year = {2025},
	note = {arXiv:2406.09211 [cs]},
}

@article{alsolami_king_2021,
	title = {King {Abdulaziz} {University} {Breast} {Cancer} {Mammogram} {Dataset} ({KAU}-{BCMD})},
	volume = {6},
	issn = {2306-5729},
	url = {https://www.mdpi.com/2306-5729/6/11/111},
	doi = {10.3390/data6110111},
	language = {en},
	number = {11},
	urldate = {2026-01-13},
	journal = {Data},
	author = {Alsolami, Asmaa S. and Shalash, Wafaa and Alsaggaf, Wafaa and Ashoor, Sawsan and Refaat, Haneen and Elmogy, Mohammed},
	month = oct,
	year = {2021},
	pages = {111},
}

@misc{bai_products-10k_2020,
	title = {Products-{10K}: {A} {Large}-scale {Product} {Recognition} {Dataset}},
	shorttitle = {Products-{10K}},
	url = {http://arxiv.org/abs/2008.10545},
	doi = {10.48550/arXiv.2008.10545},
	language = {en},
	urldate = {2026-01-13},
	publisher = {arXiv},
	author = {Bai, Yalong and Chen, Yuxiang and Yu, Wei and Wang, Linfang and Zhang, Wei},
	month = aug,
	year = {2020},
	note = {arXiv:2008.10545 [cs]},
}

@article{oquab_dinov2_nodate,
	title = {{DINOv2}: {Learning} {Robust} {Visual} {Features} without {Supervision}},
	language = {en},
	author = {Oquab, Maxime and Darcet, Timothée and Moutakanni, Théo and Vo, Huy V and Szafraniec, Marc and Khalidov, Vasil and Fernandez, Pierre and Haziza, Daniel and Massa, Francisco and El-Nouby, Alaaeldin and Assran, Mahmoud and Ballas, Nicolas and Galuba, Wojciech and Howes, Russell and Huang, Po-Yao and Li, Shang-Wen and Misra, Ishan and Rabbat, Michael and Sharma, Vasu and Synnaeve, Gabriel and Xu, Hu and Jegou, Hervé and Mairal, Julien and Labatut, Patrick and Joulin, Armand and Bojanowski, Piotr},
    year={2024}
}

@inproceedings{krause_3d_2013,
	address = {Sydney, Australia},
	title = {{3D} {Object} {Representations} for {Fine}-{Grained} {Categorization}},
	booktitle = {Proceedings of the {IEEE} {International} {Conference} on {Computer} {Vision} {Workshops} ({ICCVW})},
	author = {Krause, Jonathan and Stark, Michael and Deng, Jia and Fei-Fei, Li},
	year = {2013},
	pages = {554--561},
	doi = {10.1109/ICCVW.2013.77},
	url = {https://ieeexplore.ieee.org/document/6755874}
}

@article{sa2016deepfruits,
  title={Deepfruits: A fruit detection system using deep neural networks},
  author={Sa, Inkyu and Ge, Zongyuan and Dayoub, Feras and Upcroft, Ben and Perez, Tristan and McCool, Chris},
  journal={sensors},
  volume={16},
  number={8},
  pages={1222},
  year={2016},
  publisher={MDPI}
}

@inproceedings{sun2016benchmark,
  title={A benchmark for automatic visual classification of clinical skin disease images},
  author={Sun, Xiaoxiao and Yang, Jufeng and Sun, Ming and Wang, Kai},
  booktitle={European conference on computer vision},
  pages={206--222},
  year={2016},
  organization={Springer}
}

@article{yang2022survey,
  title={A survey of few-shot learning in smart agriculture: developments, applications, and challenges},
  author={Yang, Jiachen and Guo, Xiaolan and Li, Yang and Marinello, Francesco and Ercisli, Sezai and Zhang, Zhuo},
  journal={Plant Methods},
  volume={18},
  number={1},
  pages={28},
  year={2022},
  publisher={Springer}
}

@article{zhou2024few,
  title={Few-shot image classification of crop diseases based on vision--language models},
  author={Zhou, Yueyue and Yan, Hongping and Ding, Kun and Cai, Tingting and Zhang, Yan},
  journal={Sensors},
  volume={24},
  number={18},
  pages={6109},
  year={2024},
  publisher={MDPI}
}

\end{document}